\documentclass[journal=jacsat,manuscript=article]{achemso}

\usepackage{amsmath,amsfonts,amssymb,amsthm,dsfont}
\usepackage{times}

 \usepackage[english]{babel}
 \usepackage[T1]{fontenc}
 \usepackage[utf8]{inputenc}
 \usepackage{mathabx}
 \usepackage{booktabs}
 \usepackage{subfigure}
 \usepackage{float}
 \usepackage{wrapfig}
 \usepackage{times}
 \usepackage{textcomp}
 \usepackage{graphicx}
 \usepackage{xcolor}
 \usepackage{amsmath,amssymb}
 \usepackage{algpseudocode}
 \usepackage{algorithm}
 \usepackage{setspace}
 \mciteErrorOnUnknownfalse

\def\E{\mathbb{E}}

\usepackage[symbol]{footmisc}

\title
{Mol-CycleGAN - a generative model for molecular optimization}

\author{Łukasz Maziarka}
\affiliation{Ardigen, Kraków, Poland}
\alsoaffiliation{Jagiellonian University, Faculty of Mathematics and Computer Science, Kraków, Poland}
\email{lukasz.maziarka@ardigen.com}
\author{Agnieszka Pocha}
\affiliation{Jagiellonian University, Faculty of Mathematics and Computer Science, Kraków, Poland}
\author{Jan Kaczmarczyk}
\affiliation{Ardigen, Kraków, Poland}
\author{Krzysztof Rataj}
\affiliation{Ardigen, Kraków, Poland}
\author{Michał Warchoł}
\affiliation{Ardigen, Kraków, Poland}

\begin{document}

\maketitle

\begin{abstract}
Designing a molecule with desired properties is one of the biggest challenges in drug development, as it requires optimization of chemical compound structures with respect to many complex properties. To augment the compound design process we introduce Mol-CycleGAN -- a CycleGAN-based model that generates optimized compounds with high structural similarity to the original ones. Namely, given a molecule our model generates a structurally similar one with an optimized value of the considered property. We evaluate the performance of the model on selected optimization objectives related to structural properties (presence of halogen groups, number of aromatic rings) and to a physicochemical property (penalized logP). In the task of optimization of penalized logP of drug-like molecules our model significantly outperforms previous results.

\end{abstract}

\section{Introduction}\label{sec:intro}

The principal goal of the drug design process is to find new chemical compounds that are able to modulate the activity of a given target (typically a protein) in a desired way.\cite{ratti2001drug} However, finding such molecules in the high-dimensional chemical space of all molecules without any prior knowledge is nearly impossible. \textit{In silico} methods have been introduced to leverage the existing chemical, pharmacological and biological knowledge, thus forming a new branch of science - computer-aided drug design (CADD).\cite{srinivas2011cadd, bajorath2002cadd} Computer methods are nowadays applied at every stage of drug design pipelines \cite{srinivas2011cadd} - from the search of new, potentially active compounds,\cite{lavecchia2013vs} through optimization of their activity and physicochemical profile \cite{honorio2013h2l} and simulating their scheme of interaction with the target protein,\cite{ruyck2016docking} to assisting in planning the synthesis and evaluation of its difficulty.\cite{segler2018planning}

The recent advancements in deep learning encouraged its application in CADD.\cite{chen2018rise} The two main approaches are: \textit{virtual screening}, that is using discriminative models to screen commercial databases and classify molecules as likely active or inactive; \textit{de novo design}, that is using generative models to propose novel molecules that likely possess the desired properties. The former application already proved to give outstanding results.\cite{duvenaud2015convolutional,jastrzkebski2016learning,coley2017convolutional,pham2018graph} The latter use case is rapidly emerging, e.g. long short-term memory (LSTM) network architectures have been applied with some success.\cite{segler2017generating, bjerrum2017molecular,winter2018learning,gupta2018generative}

In the center of our interest are the hit-to-lead and lead optimization phases of the compound design process. Their goals are to optimize the drug-like molecules identified in the previous steps in terms of the desired activity profile (increased potency towards given target protein and provision of inactivity towards off-target proteins) and the physicochemical and pharmacokinetic properties. Optimizing a molecule with respect to multiple properties simultaneously remains a challenge.\cite{honorio2013h2l} Nevertheless, some successful approaches to compound generation and optimization have been proposed.

\textbf{Variational Autoencoder (VAE)} \cite{kingma2013auto} can be used for the task of molecule generation. The first such models\cite{gomez2018automatic} were based on the SMILES representation. Unfortunately, these models can generate invalid SMILES that do not correspond to any molecules. Introduction of grammars into the model improved the success rate of valid SMILES generation.\cite{kusner2017grammar,dai2018syntax} Maintaining chemical validity when generating new molecules became possible through VAEs realized directly on molecular graphs.\cite{simonovsky2018graphvae,jin2018junction}

\textbf{Generative Adversarial Networks (GAN)}\cite{goodfellow2014generative} is an alternative architecture that has been applied to \textit{de novo} drug design. GANs, together with Reinforcement Learning (RL), were recently proposed as models that generate molecules with desired properties while promoting diversity. These models use representations based on SMILES,\cite{guimaraes2017objective} graph adjacency and annotation matrices,\cite{de2018molgan} or are based on graph convolutional policy networks.\cite{you2018graph} The major obstacle in practical utility of these approaches is that the generated compounds can be difficult (or even impossible) to synthesize.

To address the problem of generating compounds difficult to synthesize, we introduce \textbf{Mol-CycleGAN} -- a generative model based on CycleGAN.\cite{zhu2017unpaired} Given a starting molecule, it generates a structurally similar one but with a desired characteristics. The similarity between these molecules is important for two reasons. First, it leads to an easier synthesis of generated molecules, and second, such optimization of the selected property is less likely to spoil the previously optimized ones, which is important in the context of multiparameter optimization. We show that our model generates molecules that possess desired properties (note that by a molecular property we also mean binding affinity towards a target protein) while retaining their structural similarity to the starting compound. Moreover, thanks to employing graph-based representation instead of SMILES, our algorithm always returns valid compounds.

We evaluate the model's ability to perform structural transformations and molecular optimization. The former indicates that the model is able to do simple structural modifications such as a change in the presence of halogen groups or number of aromatic rings. In the latter, we aim to maximize penalized logP to assess the model’s utility for compound design. Penalized logP is chosen as it is a  property often selected as a testing ground for molecule optimization models,\cite{jin2018junction,you2018graph} due to its relevance in the drug design process. In the optimization of penalized logP for drug-like molecules our model significantly outperforms previous results. To the best of our knowledge, Mol-CycleGAN is the first approach to molecule generation that uses the CycleGAN architecture.

\section{Methods}

\subsection{Junction Tree Variational Autoencoder}

JT-VAE\cite{jin2018junction} (Junction Tree Variational Autoencoder) is a method based on VAE, which works using graph structures of compounds, in contrast to previous methods which work using the SMILES representation of molecules.\cite{gomez2018automatic, kusner2017grammar,dai2018syntax}
The VAE models used for molecule generation share the encoder-decoder architecture. The encoder is a neural network used to calculate a continuous, high-dimensional representation of a molecule in the so-called latent space, whereas the decoder is another neural network used to decode a molecule from coordinates in the latent space. In VAEs the entire encoding-decoding process is stochastic (has a random component). In JT-VAE both the encoding and decoding algorithms use two components for representing the molecule: a junction-tree scaffold of molecular sub-components (called clusters) and a molecular graph.\cite{jin2018junction} JT-VAE shows superior properties compared to SMILES-based VAEs, such as 100$\%$ validity of generated molecules.

\subsection{Mol-CycleGAN}\label{sec:molgan}

Mol-CycleGAN is a novel method of performing compound optimization by learning from the \emph{sets} of molecules with and without the desired molecular property (denoted by the sets $X$ and $Y$). Our approach is to train a model to perform the transformation $G: X \rightarrow Y$ and then use this model to perform optimization of molecules. In the context of compound design $X$ ($Y$) can be, e.g., the set of inactive (active) molecules.

To represent the sets $X$ and $Y$ our approach requires an embedding of molecules which is reversible, i.e. enables both encoding and decoding of molecules.

For this purpose we use the latent space of JT-VAE, representation created by neural network during the training process. This approach has the advantage that the distance between molecules (required to calculate the loss function) can be defined directly in the latent space. Moreover, molecular properties are easier to express on graphs rather than using linear SMILES representation.\cite{weininger1988smiles} One could try formulating the CycleGAN model on the SMILES representation directly, but this would raise the problem of defining a differentiable intermolecular distance, as the standard manners of measuring similarity between molecules (Tanimoto similarity) are non-differentiable. 

\begin{figure}[t]
\centering
  \subfigure{% 
  	\centering
    \includegraphics[width=0.5\textwidth]{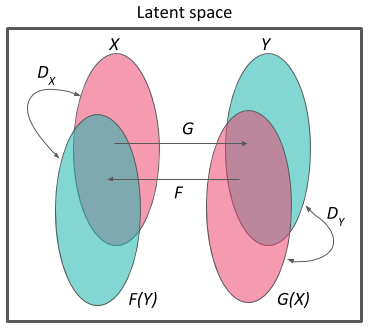} 
  }
  \caption{Schematic diagram of our Mol-CycleGAN. $X$ and $Y$ are the sets of molecules with selected values of the molecular property (e.g. active/inactive or with high/low values of logP). $G$ and $F$ are the generators. $D_X$ and $D_Y$ are the discriminators.
  }
  \label{fig:schematic}
\end{figure}

Our approach extends the CycleGAN framework \cite{zhu2017unpaired} to molecular embeddings of the latent space of JT-VAE.\cite{jin2018junction} We represent each molecule as a point in the latent space, given by the mean of the variational encoding distribution.\cite{kingma2013auto} Our model works as follows (Fig.~\ref{fig:schematic}): (i) we start by defining the sets $X$ and $Y$ (e.g., inactive/active molecules); (ii) we introduce the mapping functions  $G: X \rightarrow Y$ and $F: Y \rightarrow X$; (iii) we introduce discriminator $D_X$ (and $D_Y$) which force the generator $F$ (and $G$) to generate samples from a distribution close to the distribution of $X$ (or $Y$). The components $F$, $G$, $D_X$, and $D_Y$ are modeled by neural networks (see \textit{Workflow} for technical details). The main idea of our approach to molecule optimization is to: (i) take the prior molecule $x$ without a specified feature (e.g. specified number of aromatic rings, water solubility, activity) from set $X$, and compute its latent space embedding; (ii) use the generative neural network $G$ to obtain the embedding of molecule $G(x)$, that has this feature (as if the $G(x)$ molecule came from set $Y$) but is also similar to the original molecule $x$; (iii) decode the latent space coordinates given by $G(x)$ to obtain the SMILES representation of the optimized molecule. Thereby, the method is applicable in \textit{lead optimization} processes, as the generated compound $G(x)$ remains structurally similar to the input molecule.

\noindent To train the Mol-CycleGAN we use the following loss function:
\begin{equation}\label{eq:loss}
\begin{split}
L(G,F,D_X,D_Y) &= L_{\rm GAN}(G,D_Y,X,Y) + L_{\rm GAN}(F,D_X,Y,X)\\ &+ \lambda_1 L_{\rm cyc}(G,F) + \lambda_2 L_{\rm identity}(G,F),
\end{split}
\end{equation}
and aim to solve
$$G^*, F^* = \arg \min_{G, F} \max_{D_X, D_Y} L(G, F, D_X, D_Y).$$
We use the adversarial loss introduced in LS-GAN \cite{mao2017least}:
\begin{equation}\label{eq:loss_gan}
L_{\rm GAN}(G,D_Y,X,Y) = \frac{1}{2} \ \E_{y \sim p_{\rm data}(y)}[(D_Y(y) - 1)^2] + \frac{1}{2} \ \E_{x \sim p_{\rm data}(x)}[(D_Y(G(x)))^2],
\end{equation}
which ensures that the generator $G$ (and $F$) generates samples from a distribution close to the distribution of $Y$ (or $X$).

\noindent The cycle consistency loss 
\begin{equation}\label{eq:loss_cyc}
L_{\rm cyc}(G,F) = \E_{y \sim p_{\rm data}(y)}[\Vert G(F(y)) - y \Vert_1] +  \E_{x \sim p_{\rm data}(x)}[\Vert F(G(x)) - x \Vert_1],
\end{equation}
reduces the space of possible mapping functions, such that for a molecule $x$ from set $X$, the GAN cycle brings it back to a molecule similar to $x$, i.e. $F(G(x))$ is close to $x$ (and analogously $G(F(y))$ is close to $y$). The inclusion of the cyclic component acts as a regularization and may also help in the regime of low data, as the model can learn from both directions of the transformation. This component makes the resulting model more robust (cf. e.g. the comparison~\cite{choi2017stargan} of CycleGAN vs non-cyclic IcGAN~\cite{perarnau2016invertible}). Finally, to ensure that the generated (optimized) molecule is close to the starting one we use the identity mapping loss~\cite{zhu2017unpaired}

\begin{equation}\label{eq:loss_identity}
L_{\rm identity}(G,F) = \E_{y \sim p_{\rm data}(y)}[\Vert F(y) - y \Vert_1] +  \E_{x \sim p_{\rm data}(x)}[\Vert G(x) - x \Vert_1],
\end{equation}

which further reduces the space of possible mapping functions and prevents the model from generating molecules that lay far away from the starting molecule in the latent space of JT-VAE.

In all our experiments we use the hyperparameters $\lambda_1 = 0.3$ and $\lambda_2 = 0.1$, which were chosen by checking a couple of combinations (for structural tasks) and verifying that our optimization process: (i) improves the studied property and (ii) generates molecules similar to the starting ones. We have not performed a grid search for optimal values of $\lambda_1$ and $\lambda_2$, and hence there could be space for improvement. Note that these parameters control the balance between \textit{improvement} in the optimized property and \textit{similarity} between the generated and the starting molecule. We show in the \textit{Results} section that both the improvement and the similarity can be obtained with the proposed model.

\begin{algorithm}[t!]
  \caption{Mol-CycleGAN training algorithm. In all experiments in the paper
the following values are used: $\alpha = 0.0001$,  $\lambda_{1}=0.3$,  $\lambda_{2}=0.1$, $m=64$.} \label{algo::molcyclegan}
  \begin{algorithmic}[1] 
    \Require: $\alpha$ - the learning rate. $\lambda_{1}$ - the cycle consistency weight. $\lambda_{2}$ - the identity mapping weight. $m$, the batch size. 
    \Require: $w_X$, $w_Y$ - initial discriminators' parameters. $\theta_G$, $\theta_F$ - initial generators' parameters.
    \State Encode the $X_{\textrm{train}}, Y_{\textrm{train}}$ with the JT-VAE encoder to obtain the latent space representation of molecules from the training dataset.
    \begin{spacing}{1.4}
    \While{$\theta_G$, $\theta_F$ have not converged}  
      \State Sample $\{x^{(i)}\}_{i=1}^m \sim p(X)$ a batch from the dataset $X_{\textrm{train}}$.
      \State Sample $\{y^{(i)}\}_{i=1}^m \sim p(Y)$ a batch from the dataset $Y_{\textrm{train}}$.
      
      \State Calculate the adversarial loss. $\triangleright$ Eq.(\ref{eq:loss_gan}) \newline
      $L_{\rm GAN} \gets \frac{1}{2m} \sum_{i=1}^m \left( D_Y(y^{(i)}) - 1 \right)^{2} + \frac{1}{2m} \sum_{i=1}^m \left( D_Y(G(x^{(i)})) \right)^{2} + \frac{1}{2m} \sum_{i=1}^m \left( D_X(x^{(i)}) - 1 \right)^{2} + \frac{1}{2m} \sum_{i=1}^m \left( D_X(F(y^{(i)})) \right)^{2}$.
      
	  \State Calculate the cycle consistency loss. $\triangleright$ Eq.(\ref{eq:loss_cyc}) \newline
      $L_{\rm cyc} \gets \frac{1}{m} \sum_{i=1}^m \left|  F(G(x^{(i)})) - x^{(i)} \right| + \frac{1}{m} \sum_{i=1}^m \left|  G(F(y^{(i)})) - y^{(i)} \right|$.
      
      \State Calculate the identity mapping loss. $\triangleright$ Eq.(\ref{eq:loss_identity}) \newline 
      $L_{\rm identity} \gets \frac{1}{m} \sum_{i=1}^m \left|  G(x^{(i)}) - x^{(i)} \right| + \frac{1}{m} \sum_{i=1}^m \left|  F(y^{(i)}) - y^{(i)} \right|$.
      
      \State Calculate the loss function. $\triangleright$ Eq.(\ref{eq:loss}) \newline 
      $L \gets L_{\rm GAN} + \lambda_{1} L_{\rm cyc} + \lambda_{2} L_{\rm identity}$.
      
      \State Calculate the gradients of loss function. \newline 
      $g_G \gets \nabla_{\theta_G} L$; \ $g_F \gets \nabla_{\theta_F} L$. \newline 
      $g_X \gets \nabla_{w_X} L$; \ $g_Y \gets \nabla_{w_Y} L$.
      
      \State Minimize the loss function, using the Adam optimizer, with respect to the parameters of generators $G, F$. \newline 
      $\theta_G \gets \theta_G - \alpha \cdot \text{Adam}(\theta_G, g_G)$; \ $\theta_F \gets \theta_F - \alpha \cdot \text{Adam}(\theta_F, g_F)$. 
      
      \State Maximize the loss function, using the Adam optimizer, with respect to the parameters of discriminators $D_X, D_Y$. \newline 
      $w_X \gets w_X + \alpha \cdot \text{Adam}(w_X, g_X)$; \ $w_Y \gets w_Y + \alpha \cdot \text{Adam}(w_Y, g_Y)$. 
    \EndWhile
    \State Decode $G(X_{\textrm{test}})$ and $F(Y_{\textrm{test}})$ with the JT-VAE decoder to obtain the SMILES representation of molecules returned by the Mol-CycleGAN.
    \end{spacing}
\end{algorithmic}
\end{algorithm}

\subsection{Workflow}

We conduct experiments to test whether the proposed model is able to generate molecules that possess desired properties and are close to the starting molecules. 
Namely, we evaluate the model on tasks related to \textit{structural modifications}, as well as on tasks related to \textit{molecule optimization}. For testing molecule optimization we select the octanol-water partition coefficient (logP) penalized by the synthetic accessibility (SA) score. logP describes lipophilicity - a parameter influencing a whole set of other characteristics of compounds such as solubility, permeability through biological membranes, ADME (absorption, distribution, metabolism, and excretion) properties, and toxicity. We use the formulation as reported in the paper on JT-VAE,\cite{jin2018junction} i.e. for molecule $m$ the penalized logP is given as $logP(m)-SA(m)$. We use the ZINC-250K dataset used in similar studies\cite{kusner2017grammar, jin2018junction} which contains 250 000 drug-like molecules extracted from the ZINC database.\cite{sterling2015zinc} The detailed formulation of the tasks is the following:
\begin{itemize}
\item{\textbf{Structural transformations}} We test the model's ability to perform simple structural transformations of the molecules. To this end, we choose the sets $X$ and $Y$, differing in some structural aspects, and then test if our model can learn the transformation rules and apply them to molecules previously unseen by the model. There are two features by which we divide the sets:
\begin{itemize}
\item \textbf{Halogen moieties} We split the dataset into two subsets $X$ and $Y$. The set $Y$ consists of molecules which contain at least one of the following SMARTS: '[!\#1]Cl', '[!\#1]F', '[!\#1]I', 'C\#N', whereas the set $X$ consists of such molecules which do not contain any of them. The SMARTS chosen in this experiment indicate halogen moieties and the nitrile group. Their presence and position within a molecule can have an immense impact on the compound’s activity. 

\item \textbf{Aromatic rings} Molecules in $X$ have exactly two aromatic rings, whereas molecules in $Y$ have one or three aromatic rings.
\end{itemize}

\item \textbf{Constrained molecule optimization}
We optimize penalized logP, while constraining the degree of deviation from the starting molecule. The similarity between molecules is measured with Tanimoto similarity on Morgan Fingerprints.\cite{rogers2010extended} The sets $X$ and $Y$ are random samples from ZINC-250K, where the compounds' penalized logP values are below and above the median, respectively.

\item \textbf{Unconstrained molecule optimization}
We perform unconstrained optimization of penalized logP. The set $X$ is a random sample from ZINC-250K and the set $Y$ is a random sample from the top-20$\%$ molecules with the highest penalized logP in ZINC-250K.

\end{itemize}

\subsection{Composition of the datasets}

\paragraph{Dataset sizes}

In Table~\ref{table:B} we show the number of molecules in the datasets used for training and testing. In all experiments we use separate sets for training the model ($X_{\textrm{train}}$ and $Y_{\textrm{train}}$) and separate, non-overlapping ones for evaluating the model ($X_{\textrm{test}}$ and $Y_{\textrm{test}}$). In constrained and unconstrained molecular optimization experiments no $Y_{\textrm{test}}$ set is required.

\begin{table}[H]
\caption{Dataset sizes}
\centering
\begin{tabular}{lcccc}
\toprule
Dataset &  Halogen &  Aromatic & Constrained & Unconstrained\\ & moieties & rings &  optimization & optimization\\
\midrule
 $X_{\textrm{train}}$ &             75000 &             80000 &          80000 &          80000 \\
  $X_{\textrm{test}}$ &             86899 &             18220 &            800 &            800 \\
\midrule
 $Y_{\textrm{train}}$ &             75000 &             80000 &          80000 &          24946 \\
  $Y_{\textrm{test}}$ &             12556 &             43193 &              - &              - \\
\bottomrule
\end{tabular}
\label{table:B}
\end{table}

\paragraph{Distribution of the selected properties}

In the experiment on halogen moieties, the set $X$ always (i.e., both in train- and test-time) contains molecules without halogen moieties, and the set $Y$ always contains molecules with halogen moieties. In the dataset used to construct the latent space (ZINC-250K) 65 \% molecules do not contain any halogen moiety, whereas the remaining 35 \% contain one or more halogen moieties.

\begin{figure}[H]
\centering
    \includegraphics[width=0.7\textwidth]{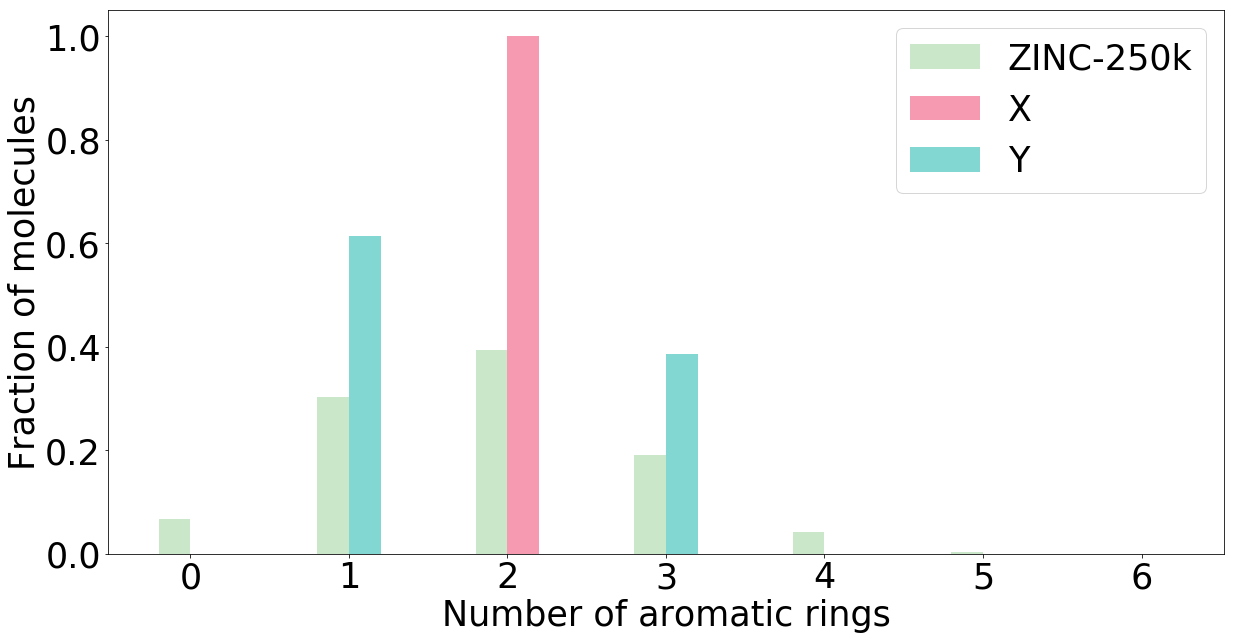}\label{fig:prior_412}
\caption{Number of aromatic rings in ZINC-250K and in the sets used in the experiment on aromatic rings.}
\label{figB1}
\end{figure}

In the experiment on aromatic rings, the set $X$ always (i.e., both in train- and test-time) contains molecules with 2 rings, and the set $Y$ always contains molecules with 1 or 3 rings. The distribution of the number of aromatic rings in the dataset used to construct the latent space (ZINC-250K) is shown in Figure~\ref{figB1} along with the distribution for $X$ and $Y$. 

\begin{figure}[H]
\centering
    \includegraphics[width=0.7\textwidth]{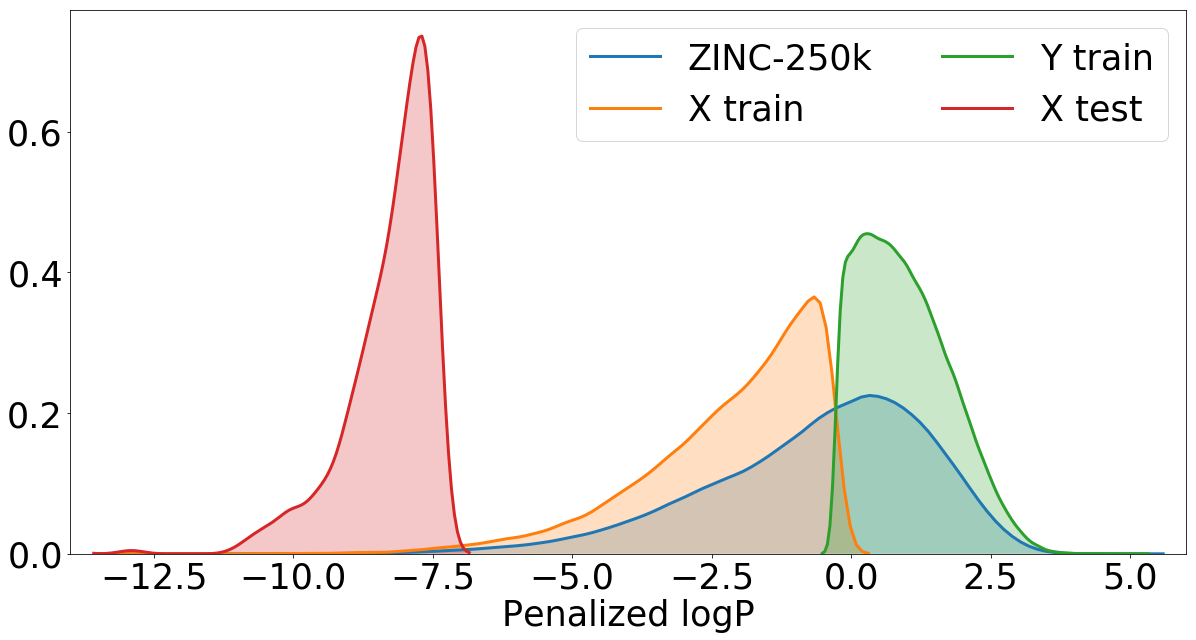}\label{fig:prior_42}
\caption{Distribution of penalized logP in ZINC-250K and in the sets used in the task of constrained molecule optimization. Note that the sets $X_{\textrm{train}}$ and $Y_{\textrm{train}}$  are non-overlapping (they are a random sample from ZINC-250K split by the median). $X_{\textrm{test}}$ is the set of 800 molecules from ZINC-250K with the lowest values of penalized logP.} 
\label{figB2}
\end{figure}

\begin{figure}[H]
\centering
    \includegraphics[width=0.7\textwidth]{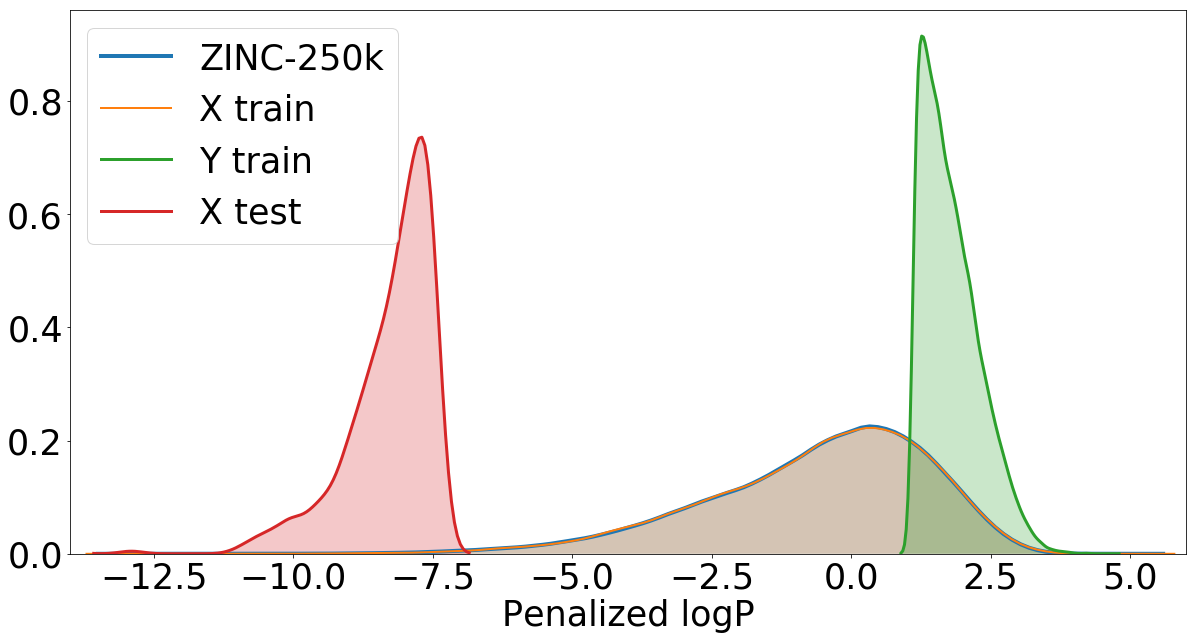}\label{fig:prior_43}
\caption{Distribution of penalized logP in ZINC-250K and in the sets used in the task of unconstrained molecule optimization. Note that the set $X_{\textrm{train}}$ is a random sample from ZINC-250K, and hence the same distribution is observed for the two sets.} 
\label{figB3}
\end{figure}

For the molecule optimization tasks we plot the distribution of the property being optimized (penalized logP) in Figs.~\ref{figB2} (constrained optimization) and \ref{figB3} (unconstrained optimization). 

\subsection{Architecture of the models}
\label{app:models}

All networks are trained using the Adam optimizer \cite{kingma2014adam} with learning rate $0.0001$. During training we use batch normalization.\cite{ioffe2015batch} As the activation function we use leaky-ReLU with $\alpha = 0.1$. In the structural experiments the models are trained for 100 epochs and in the physiochemical experiments for 300 epochs.

\subsubsection{Structural data experiments}

\begin{itemize}
\item \textbf{Generators} are built of one fully connected residual layer, followed by one dense layer. All layers contain 56 units.
\item \textbf{Discriminators} are built of 6 dense layers of the following sizes: 56, 42, 28, 14, 7, 1 units.
\end{itemize}

\subsubsection{Physiochemical data experiments}

\begin{itemize}
\item \textbf{Generators} are built of four fully connected residual layers. All layers contain 56 units.
\item \textbf{Discriminators} are built of 7 dense layers of the following sizes: 48, 36, 28, 18, 12, 7, 1 units.
\end{itemize}

\section{Results}\label{sec:results}

\subsection{Structural transformations}\label{sect:4.1}

In each structural experiment we test the model's ability to perform simple transformations of molecules in both directions $X \rightarrow Y$ and $Y \rightarrow X$. Here, $X$ and $Y$ are non-overlapping sets of molecules with a specific structural property. We start with experiments on structural properties because they are easier to interpret and the rules related to transforming between $X$ and $Y$ are well defined.  Hence, the present task should be easier for the model, as compared to the optimization of complex molecular properties, for which there are no simple rules connecting $X$ and $Y$.

\begin{table}[h]
\caption {Evaluation of models modifying the presence of halogen moieties and the number of aromatic rings. Success rate is the fraction of times when a desired modification occurs. Non-identity is the fraction of times when the generated molecule is different from the starting one. Uniqueness is the fraction of unique molecules in the set of generated molecules.}
\label{tab:success_rate}
\begin{center}
\begin{tabular}{l | cc | cc}
\toprule
 & \multicolumn{2}{c |}{Halogen moieties} & \multicolumn{2}{c}{Aromatic rings} \\
                 & $X \rightarrow G(X)$ & $Y \rightarrow F(Y)$ &      $X \rightarrow G(X)$ & $Y \rightarrow F(Y)$ \\
\midrule
    Success rate &    0.6429 &  0.7161 &  0.5342 &  0.4216 \\
    Non-identity &    0.9345 &  0.9574 &  0.9082 &  0.8899 \\
      Uniqueness &    0.9952 &  0.9953 &  0.9957 &  0.9954 \\
\bottomrule
\end{tabular}
\end{center}
\end{table}

\begin{figure}[t]
\centering
  \subfigure{% 
  	\centering
    \includegraphics[width=0.49\textwidth]{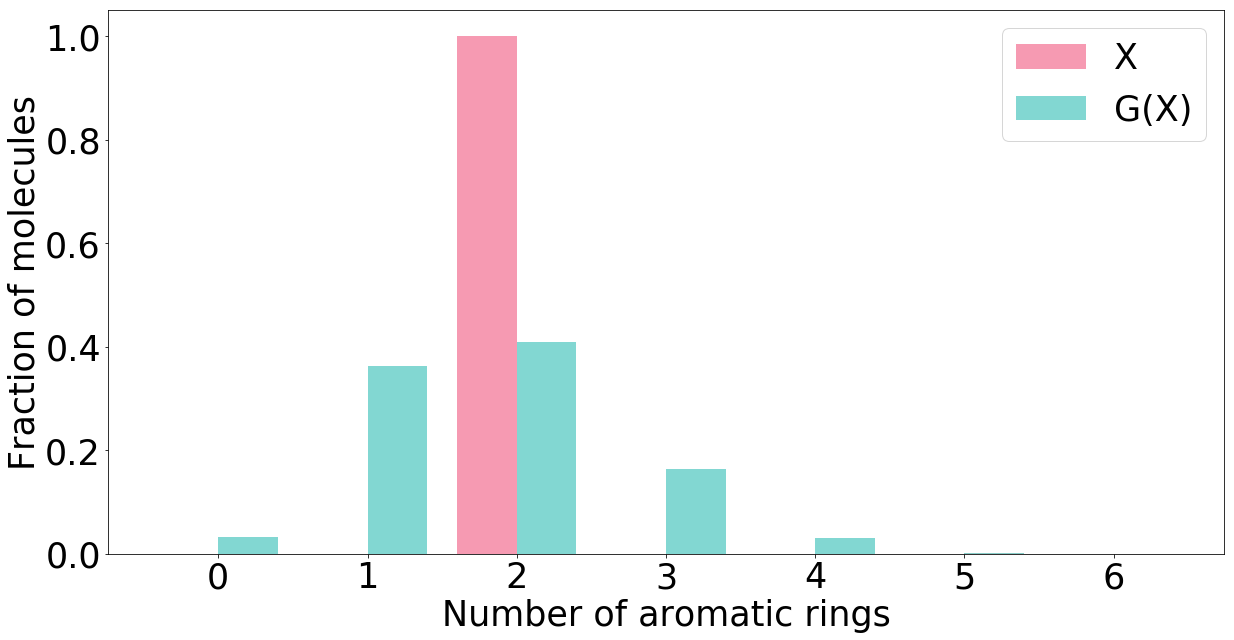} 
  }
  \subfigure{% 
    \centering
    \includegraphics[width=0.49\textwidth]{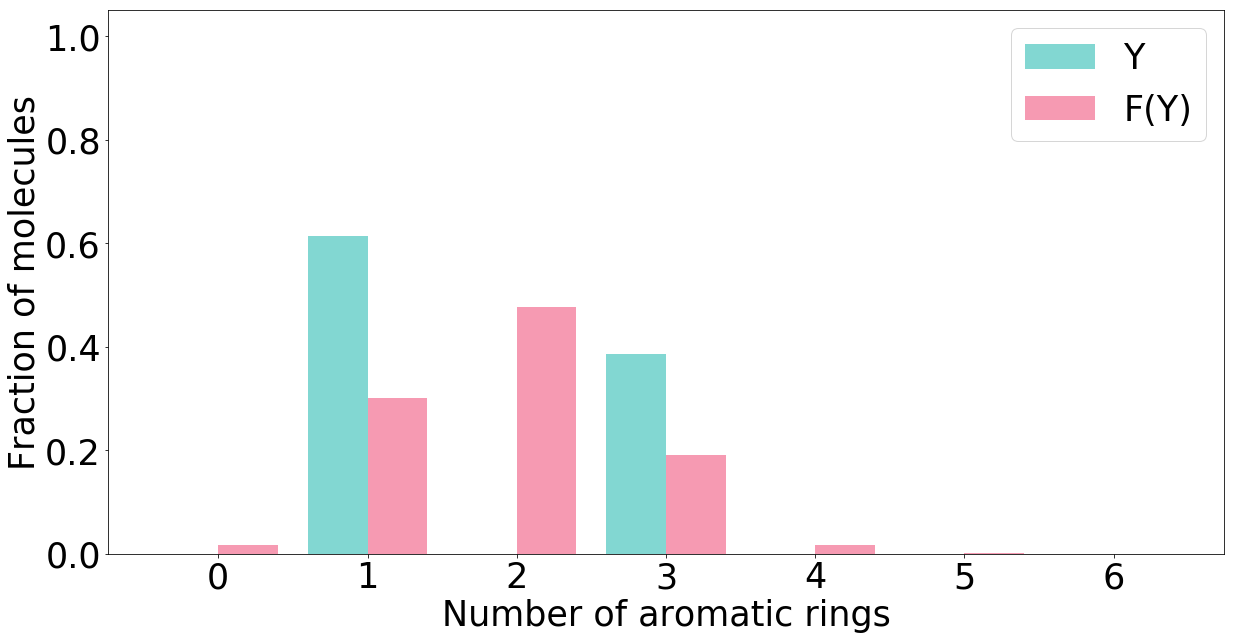} 
  } 
  \caption{Distributions of the number of aromatic rings in $X$ and $G(X)$ (left), and $Y$ and $F(Y)$ (right). Identity mappings are not included in the figures.}
  \label{fig:aromatic_rings}
\end{figure}

In Table \ref{tab:success_rate} we show the success rates for the tasks of performing structural transformations of molecules. The task of changing the number of aromatic rings is more difficult than changing the presence of halogen moieties. In the former the transition between $X$ (with 2 rings) and $Y$ (with 1 or 3 rings, cf. Fig. \ref{fig:aromatic_rings}) is more than a simple addition/removal transformation, as it is in the other case (see Fig. 5 for the distributions of the aromatic rings). This is reflected in the success rates which are higher for the task of transformations of halogen moieties. In the dataset used to construct the latent space (ZINC-250K) 64.9 \% molecules do not contain any halogen moiety, whereas the remaining 35.1 \% contain one or more halogen moieties. This imbalance might be the reason for the higher success rate in the task of removing halogen moieties ($Y \rightarrow F(Y)$). Molecular similarity and drug-likeness are achieved in all experiments.

\begin{figure}[t]
\centering
  \subfigure[Halogen moieties]{% 
  	\centering
    \includegraphics[width=0.49\textwidth]{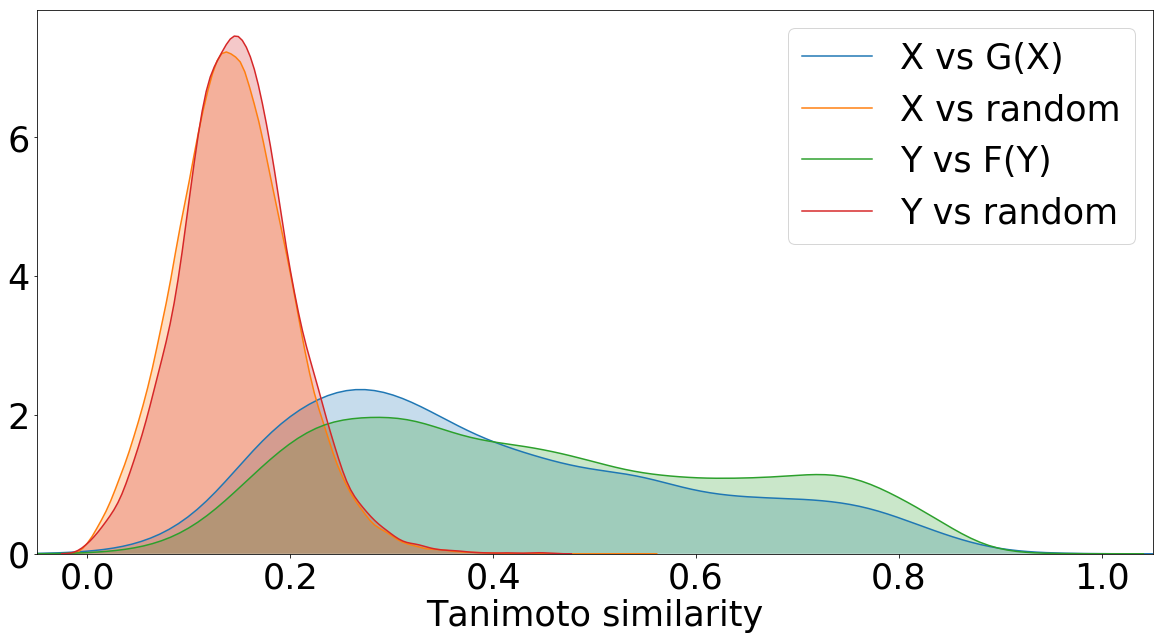}\label{fig:halogenes_1_0_tanimoto} 
  } 
  \subfigure[Aromatic rings]{% 
    \centering
    \includegraphics[width=0.49\textwidth]{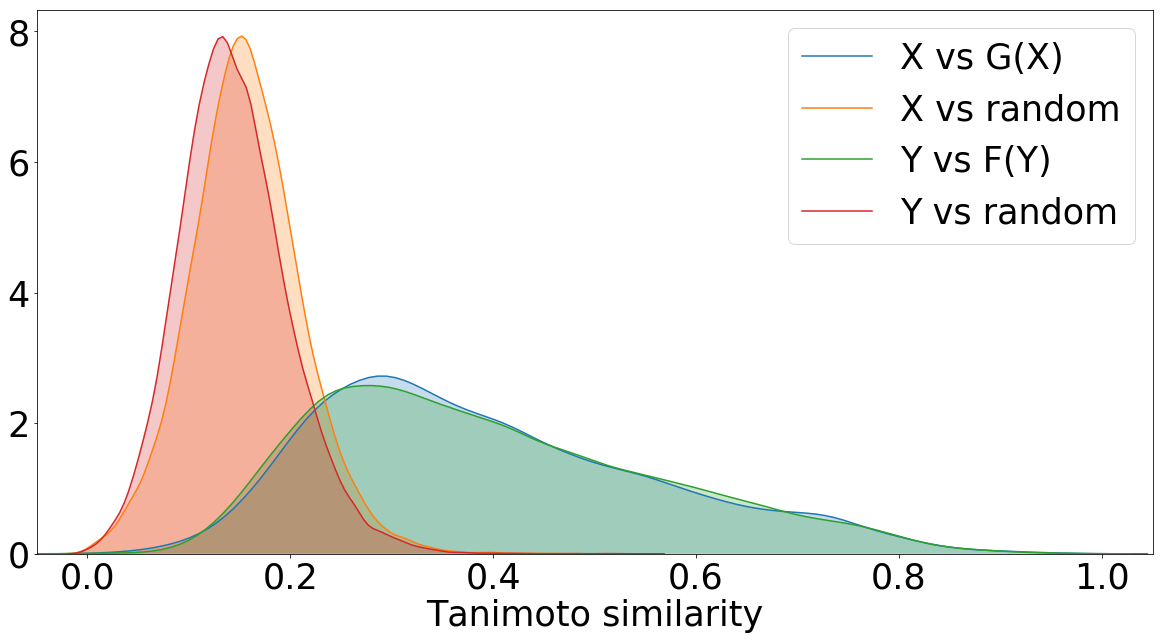}\label{fig:aromatic_rings_13_2_tanimoto} 
  }
  \caption{Density plots of Tanimoto similarities between molecules from $Y$ (and $X$) and their corresponding molecules from $F(Y)$ (and $G(X)$). Similarities between molecules from $Y$ (and $X$) and random molecules from ZINC-250K are included for comparison. Identity mappings are not included. The distributions of similarities related to transformations given by $G$ and $F$ show the same trend.} 
  \label{fig:aromatic_rings_tanimoto}
\end{figure}

\begin{figure}[t]
\centering
  \subfigure[top: $X$; bottom: $G(X)$]{% 
  	\centering
    \includegraphics[width=0.49\textwidth]{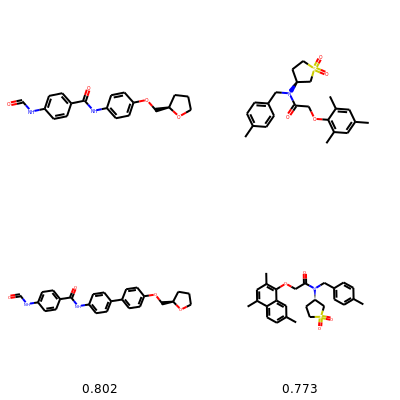}\label{fig:aromatic_rings_2_13_most_similar} 
  } 
  \subfigure[top: $Y$; bottom: $F(Y)$]{% 
    \centering
    \includegraphics[width=0.49\textwidth]{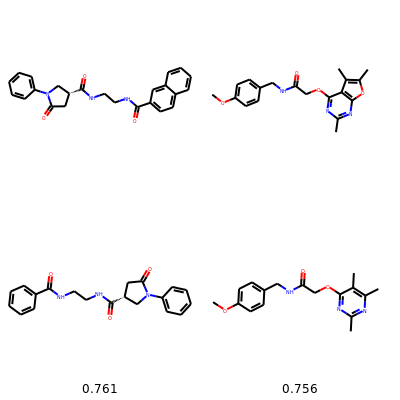}\label{fig:aromatic_rings_13_2_most_similar} 
  }
  \caption{The most similar molecules with changed number of aromatic rings. In the top row we show the starting molecules, whereas in the bottom row we show the generated molecules. Below we provide the Tanimoto similarities between the molecules.} 
  \label{fig:aromatic_rings_most_similar}
\end{figure}

To confirm that the generated molecules are close to the starting ones, we show in Figure~\ref{fig:aromatic_rings_tanimoto} distributions of their Tanimoto similarities (using Morgan fingerprints). For comparison we also include distributions of the Tanimoto similarities between the starting molecule and a random molecule from the ZINC-250K dataset. The high similarities between the generated and the starting molecules show that our procedure is neither a random sampling from the latent space, nor a memorization of the manifold in the latent space with the desired value of the property. In Figure~\ref{fig:aromatic_rings_most_similar} we visualize the molecules, which after transformation are the most similar to the starting molecules.

	\subsection{Constrained molecule optimization}\label{sect:4.2}

As our main task we optimize the desired property under the constraint that the similarity between the original and the generated molecule is higher than a fixed threshold. This is a more realistic scenario in drug discovery, where the development of new drugs usually starts with known molecules such as existing drugs.\cite{besnard2012automated} Here, we maximize the penalized logP coefficient and use the Tanimoto similarity with the Morgan fingerprint\cite{rogers2010extended} to define the threshold of similarity, $sim(m, m') \geq \delta$. We compare our results with previous similar studies.\cite{jin2018junction,you2018graph}

In our optimization procedure each molecule (given by the latent space coordinates $x$) is fed into the generator to obtain the `optimized' molecule $G(x)$. The pair $(x, G(x))$ defines what we call an 'optimization path' in the latent space of JT-VAE. To be able to make a comparison with the previous research~\cite{jin2018junction} we start the procedure from the 800 molecules with the lowest values of penalized logP in ZINC-250K and then we decode molecules from $K = 80$ points along the path from $x$ to $G(x)$ in equal steps.

From the resulting set of molecules we report the molecule with the highest penalized logP score that satisfies the similarity constraint. A modification succeeds if one of the decoded molecules satisfies the constraint and is distinct from the starting one.

\begin{table}[t]
\caption{Results of the constrained optimization for JT-VAE~\cite{jin2018junction}, GCPN~\cite{you2018graph} and Mol-CycleGAN.}
\label{table:constrained}
\begin{center}
\begin{tabular}{c | cc | cc | cc }
\toprule
 & \multicolumn{2}{c}{JT-VAE} & \multicolumn{2}{c}{GCPN} & \multicolumn{2}{c}{Mol-CycleGAN} \\
 Delta &    Improvement &     Similarity &    Improvement &     Similarity &    Improvement &     Similarity \\
\midrule
     0 &  1.91 $\pm$ 2.04 &  0.28 $\pm$ 0.15 &   4.20 $\pm$ 1.28 & 0.32 $\pm$ 0.12 &   \textbf{8.30} $\pm$ 1.98 &  0.16 $\pm$ 0.09  \\
   0.2 &  1.68 $\pm$ 1.85 & 0.33 $\pm$ 0.13 &   4.12 $\pm$ 1.19 &  0.34 $\pm$ 0.11 &  \textbf{5.79} $\pm$ 2.35 &   0.30 $\pm$ 0.11  \\
   0.4 &  0.84 $\pm$ 1.45 &  0.51 $\pm$ 0.10 &  2.49 $\pm$ 1.30 &   0.47 $\pm$ 0.08 &   \textbf{2.89} $\pm$ 2.08 &   0.52 $\pm$ 0.10  \\
   0.6 &  0.21 $\pm$ 0.75 &  0.69 $\pm$ 0.06 &  0.79 $\pm$ 0.63 &  0.68 $\pm$ 0.08 &  \textbf{1.22} $\pm$ 1.48 &  0.69 $\pm$ 0.07  \\
\bottomrule
\end{tabular}
\end{center}
\end{table}

\begin{table}[t]
\caption{Success rate for constrained optimization for JT-VAE~\cite{jin2018junction}, GCPN~\cite{you2018graph} and Mol-CycleGAN.}
\label{table:constrained_success}
\begin{center}
\begin{tabular}{c | c | c | c }
\toprule
 & JT-VAE & GCPN & Mol-CycleGAN \\
\midrule
     0 &  97.5\% & \textbf{100.0}\% &  99.75\% \\
   0.2 &  97.1\% & \textbf{100.0}\% &  93.75\% \\
   0.4 &  83.6\% & \textbf{100.0}\% &  58.75\% \\
   0.6 &  46.4\% & \textbf{100.0}\% &  19.25\% \\
\bottomrule
\end{tabular}
\end{center}
\end{table}

 In the task of optimizing penalized logP of \emph{drug-like} molecules, our method significantly outperforms the previous results in the mean improvement of the property (see Table~\ref{table:constrained}). It achieves a comparable mean similarity in the constrained scenario (for $\delta > 0$). The success rates are comparable for $\delta = 0, 0.2$, whereas for the more stringent constraints ($\delta = 0.4, 0.6$) our model has lower success rates (see Table~\ref{table:constrained_success}). Note that comparably high improvements of penalized logP can be obtained using reinforcement learning.\cite{you2018graph} However, the molecules optimized in such a manner are not drug-like, e.g., they have very low quantitative estimate of drug-likeness score~\cite{bickerton2012quantifying} even for $\delta > 0$.  In our method (as well as in JT-VAE) drug-likeness is achieved `by construction' and is an intrinsic feature of the latent space obtained by training the variational autoencoder on molecules from ZINC (which are drug-like).

\begin{figure}[t]
\centering
    \includegraphics[width=0.98\textwidth]{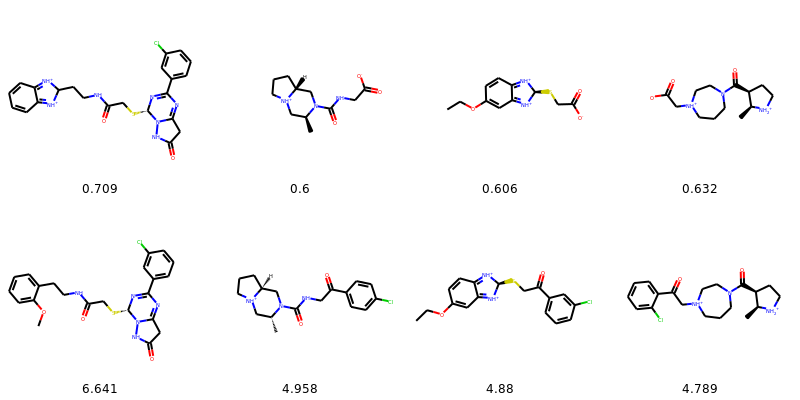}\label{fig:constrained_logP_improvements_similar_gradient_ascent}
\caption{Molecules with the highest improvement of the penalized logP for $\delta \ge 0.6$. In the top row we show the starting molecules, whereas in the bottom row we show the optimized molecules. Upper row numbers indicate Tanimoto similarities between the starting and the final molecule. The improvement in the score is given below the generated molecules.} 
\end{figure}

\subsubsection{Molecular paths from constrained optimization experiments}

\begin{figure}[H]
\centering
    \includegraphics[width=0.8\textwidth]{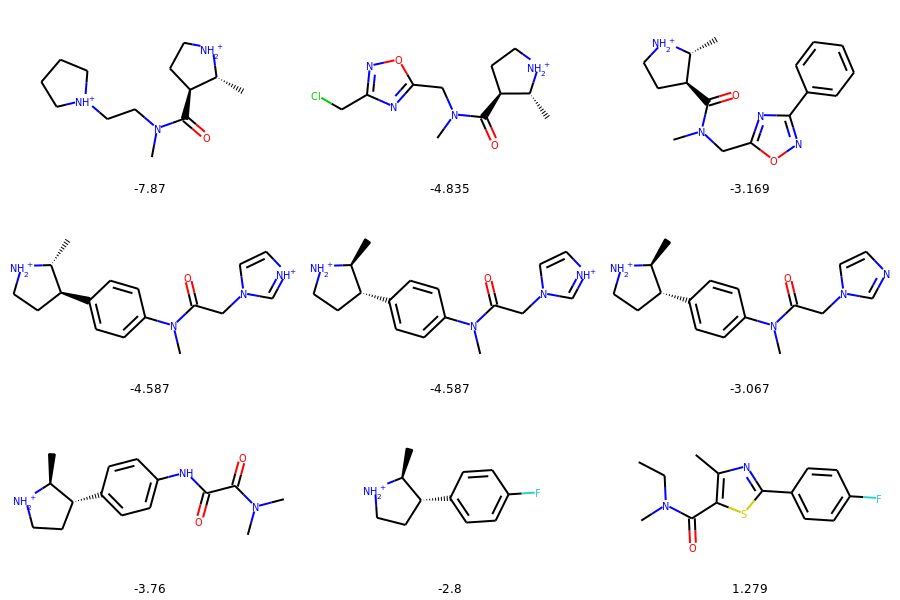}\label{fig:walk_constrained_2}
\caption{Evolution of a selected exemplary molecule during constrained optimization. We only include the steps along the path where a change in the molecule is introduced. We show values of penalized logP below the molecules.} 
\label{fig:constrained:paths1}
\end{figure}

\begin{figure}[H]
\centering
    \includegraphics[width=0.8\textwidth]{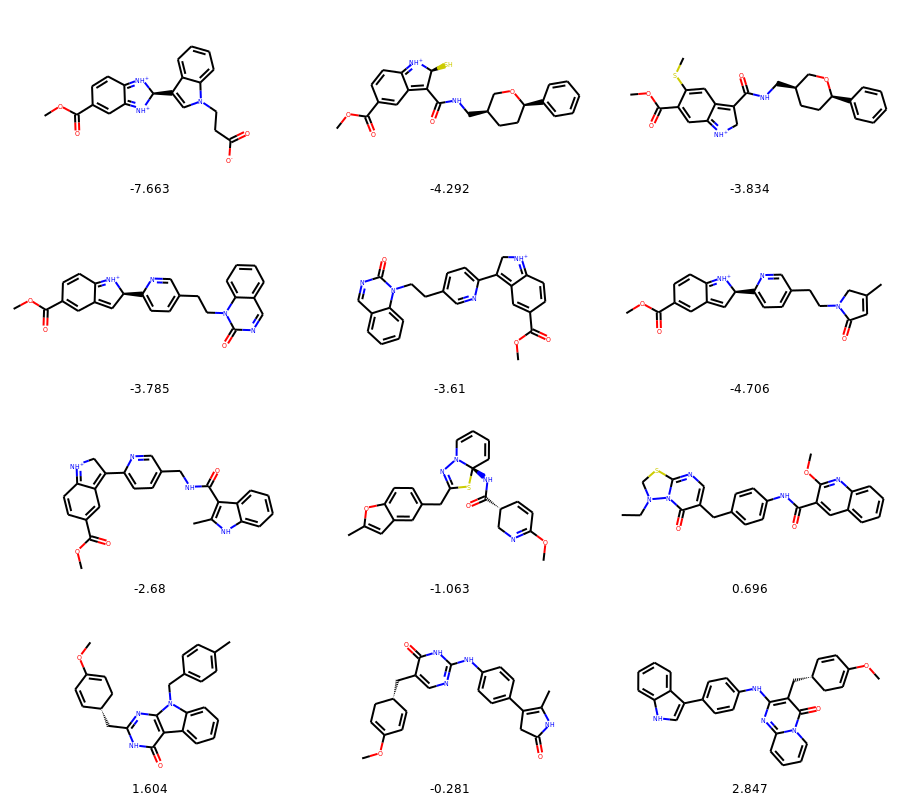}\label{fig:walk_constrained_3}
\caption{Evolution of a selected exemplary molecule during constrained optimization. We only include the steps along the path where a change in the molecule is introduced. We show values of penalized logP below the molecules.} 
\label{fig:constrained:paths2}
\end{figure}

\begin{figure}[H]
\centering
    \includegraphics[width=0.8\textwidth]{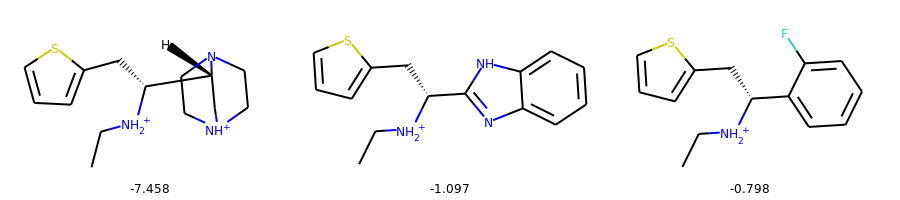}\label{fig:walk_constrained_1}
\caption{Evolution of a selected exemplary molecule during constrained optimization. We only include the steps along the path where a change in the molecule is introduced. We show values of penalized logP below the molecules.} 
\label{fig:constrained:paths3}
\end{figure}

In the following section we show examples of the evolution of the selected molecules for the constrained optimization experiments. Figures~\ref{fig:constrained:paths1}-\ref{fig:constrained:paths3} show starting and final molecules, together with all molecules generated along the optimization path and their values of penalized logP. 

\subsection{Unconstrained molecule optimization}\label{sect:4.3}

Our architecture is tailor made for the scenario of constrained molecule optimization. However, as an additional task, we check what happens when we iteratively use the generator on the molecules being optimized. This should lead to diminishing similarity between the starting molecules and those in consecutive iterations. For the present task the set $X$ needs to be a sample from the entire ZINC-250K, whereas the set $Y$ is chosen as a sample from the top-20$\%$ of molecules with the highest value of penalized logP. Each molecule is fed into the generator and the corresponding `optimized' molecule's latent space representation is obtained. The generated latent space representation is then treated as the new input for the generator. The process is repeated $K$ times and the resulting set of molecules is $\{G(x), G(G(x)),$ ... \}. Here, as in the previous task and as in previous research~\cite{jin2018junction} we start the procedure from the 800 molecules with the lowest values of penalized logP in ZINC-250K.

\begin{figure}[t]
\centering
  \subfigure[]{% 
  	\centering
    \includegraphics[height=0.24\textwidth]{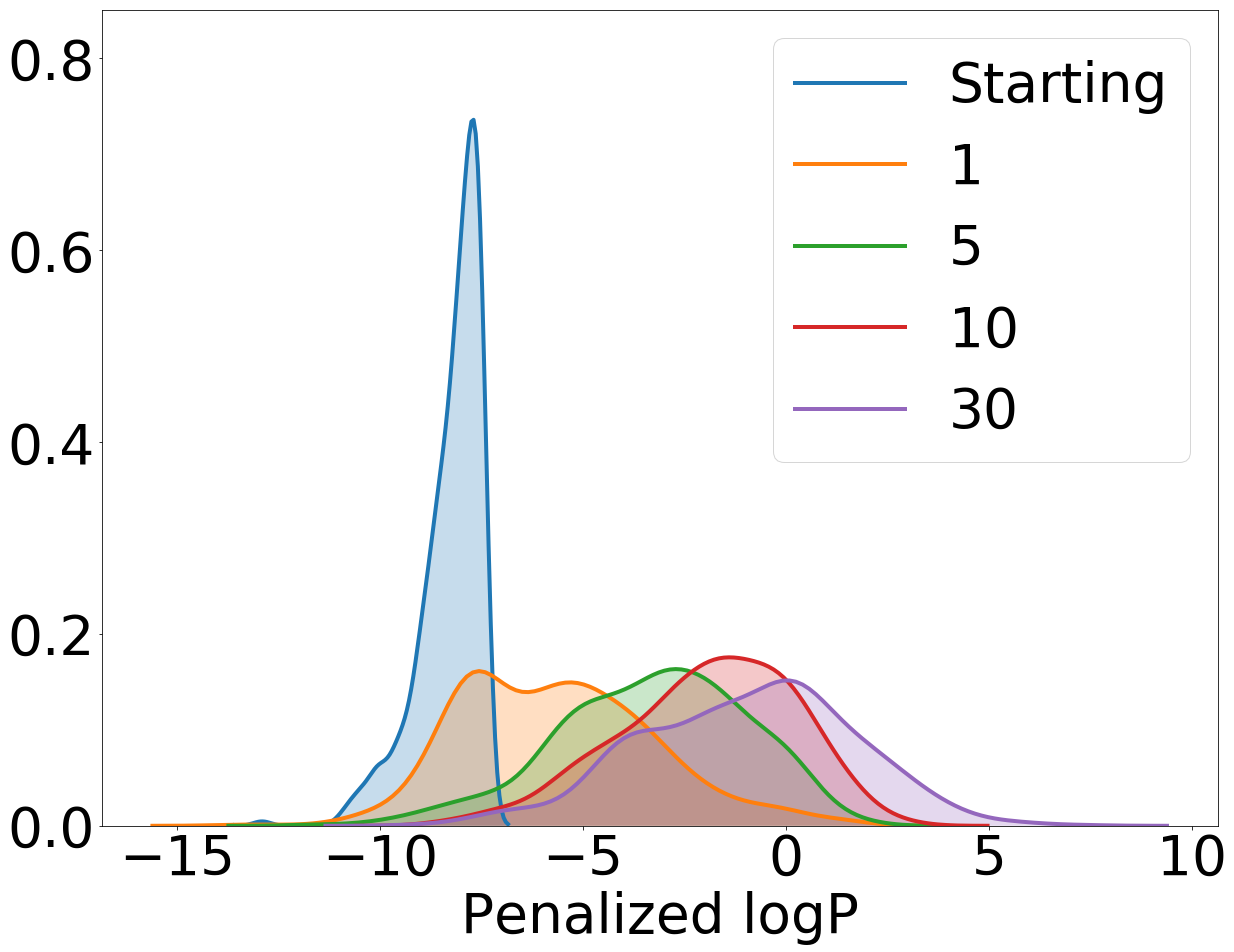}\label{fig:optimization_density_iters} 
  } 
  \subfigure[]{% 
  	\centering
    \includegraphics[height=0.24\textwidth]{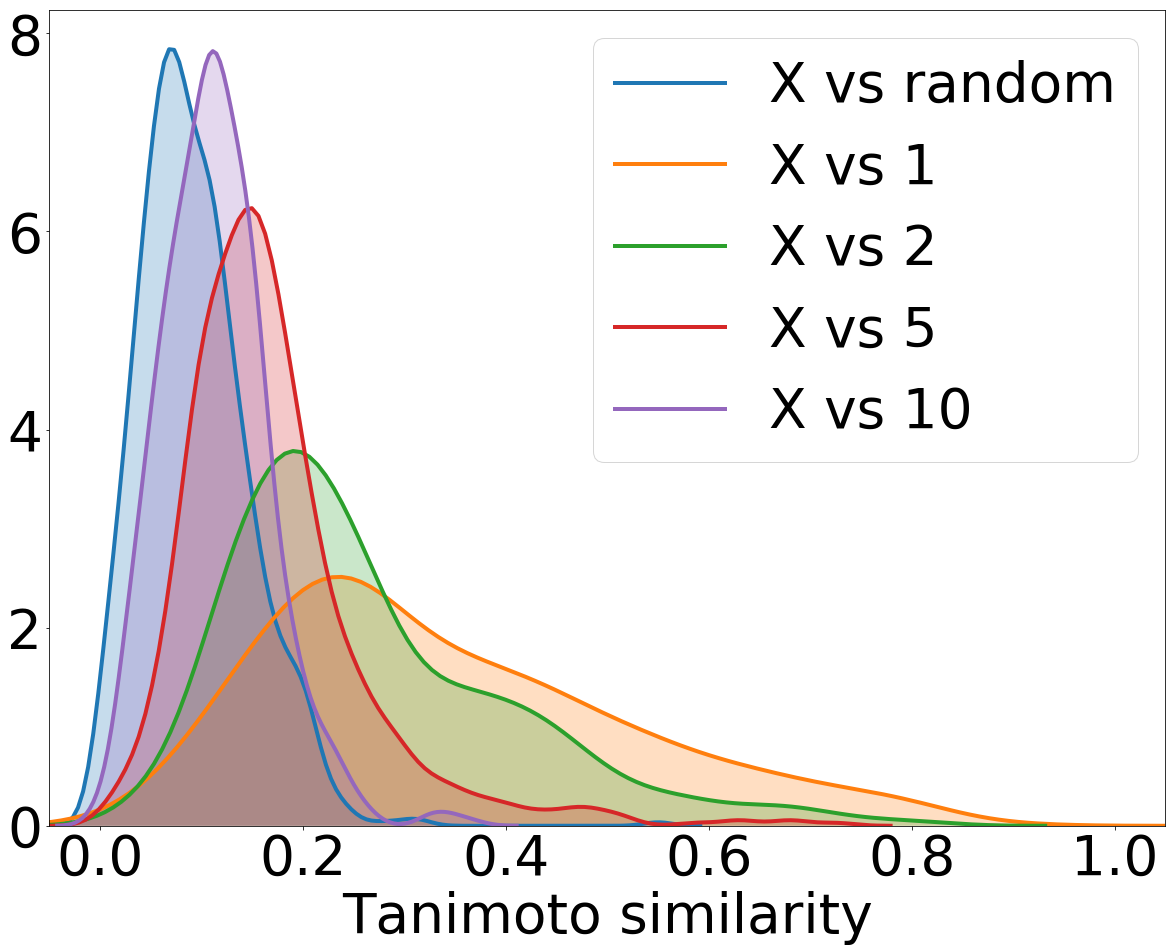}\label{fig:optimization_tanimoto} 
  } 
  \subfigure[]{% 
    \centering
    \includegraphics[height=0.24\textwidth]{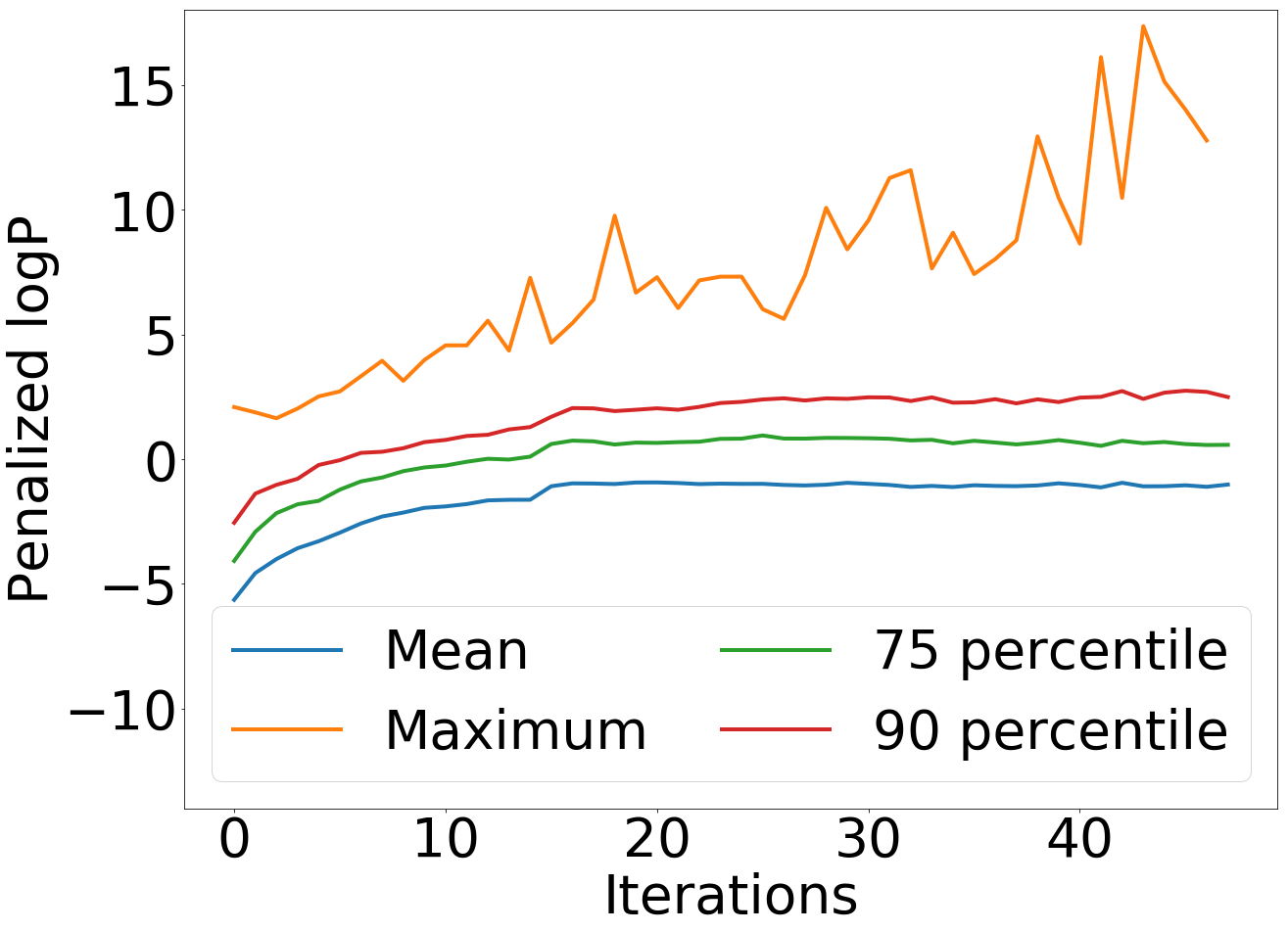}\label{fig:optimization_means_iters} 
  }
  \caption{Results of iterative procedure of the unconstrained optimization. (a) Distribution of the penalized logP in the starting set and after $K=1, 5, 10, 30$ iterations. (b) Distribution of the Tanimoto similarity between the starting molecules $X$ and random molecules from ZINC-250K, as well as those generated after $K=1, 2, 5, 10$ iterations. (c) Plot of the mean value, percentiles (75th and 90th), and the maximum value of penalized logP as a function of the number of iterations.} 
  \label{fig:unconstrained_opt}
\end{figure}

The results of our unconstrained molecule optimization are shown in Figure~\ref{fig:unconstrained_opt}. In Figure~\ref{fig:unconstrained_opt}(a) and (c) we observe that consecutive iterations keep shifting the distribution of the objective (penalized logP) towards higher values. However, the improvement from further iterations is decreasing. Interestingly, the maximum of the distribution keeps increasing (although in somewhat random fashion). After 10-20 iterations it reaches very high values of logP observed from molecules which are not drug-like, similarly to those obtained with RL.~\cite{you2018graph} Both in the case of the RL approach and in our case, the molecules with the highest penalized logP after many iterations also become non-drug-like -- see Figure~\ref{figC1} for a list of compounds with the maximum values of penalized logP in the iterative optimization procedure. This lack of drug-likeness is related to the fact that after performing many iterations, the distribution of coordinates of our set of molecules in the latent space goes far away from the prior distribution (multivariate normal) used when training the JT-VAE on ZINC-250K. 
In Fig.~\ref{fig:unconstrained_opt}(b) we show the evolution of the distribution of Tanimoto similarities between the starting molecules and those obtained after $K = 1, 2, 5, 10$ iterations. We also show the similarity between the starting molecules and random molecules from ZINC-250K. We observe that after 10 iterations the similarity between the starting molecules and the optimized ones is comparable to the similarity of random molecules from ZINC-250K. After around 20 iterations the optimized molecules become less similar to the starting ones than random molecules from ZINC-250K, as the set of optimized molecules is moving further away from the space of drug-like molecules.

\subsubsection{Molecular paths from unconstrained optimization experiments}

In the following section we show examples of the evolution of selected molecules for the unconstrained optimization experiments. Figures 13 and 14 show starting and final molecules, together with all molecules generated during the iteration over the optimization path and their penalized logP values.

\begin{figure}[H]
\centering
    \includegraphics[width=0.8\textwidth]{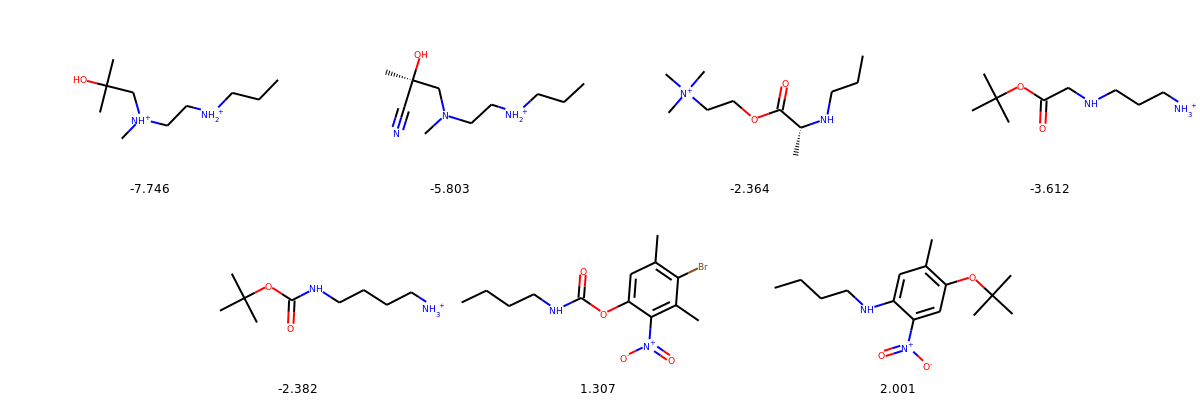}\label{fig:walk_unconstrained_1}
\caption{Evolution of a selected molecule during consecutive iterations of unconstrained optimization. We show values of penalized logP below the molecules.} 
\end{figure}

\begin{figure}[H]
\centering
    \includegraphics[width=0.8\textwidth]{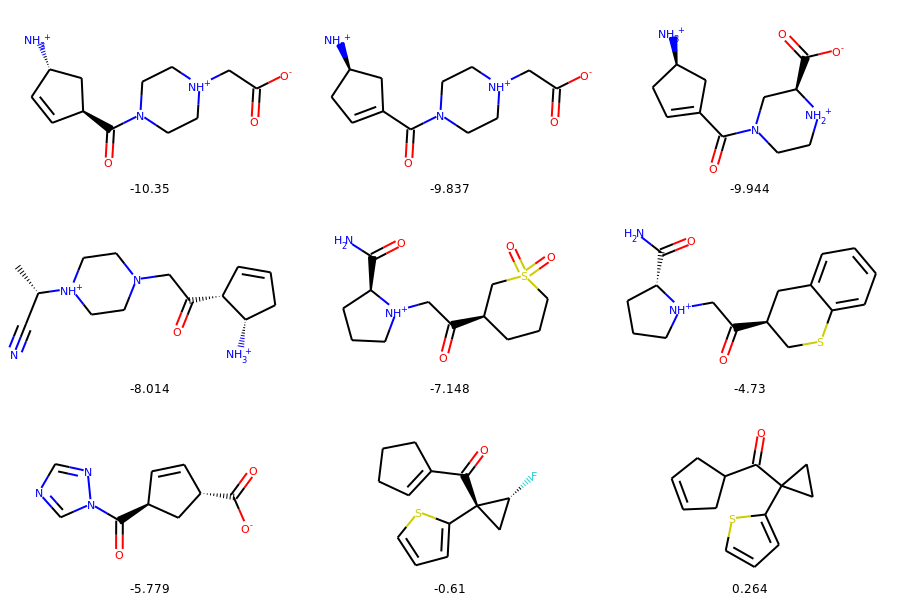}\label{fig:walk_unconstrained_2}
\caption{Evolution of a selected molecule during consecutive iterations of unconstrained optimization. We show values of penalized logP below the molecules.} 
\end{figure}

\subsubsection{Molecules with the highest values of penalized logP}

On the Figure 12(c) we plot the maximum value of penalized logP in the set of molecules being optimized, as a function of number of iterations of for unconstrained molecule optimization. In Figure 15 we show corresponding molecules for iterations 1-24.

\begin{figure}[H]
\centering
    \includegraphics[width=\textwidth,height=0.88\textheight]{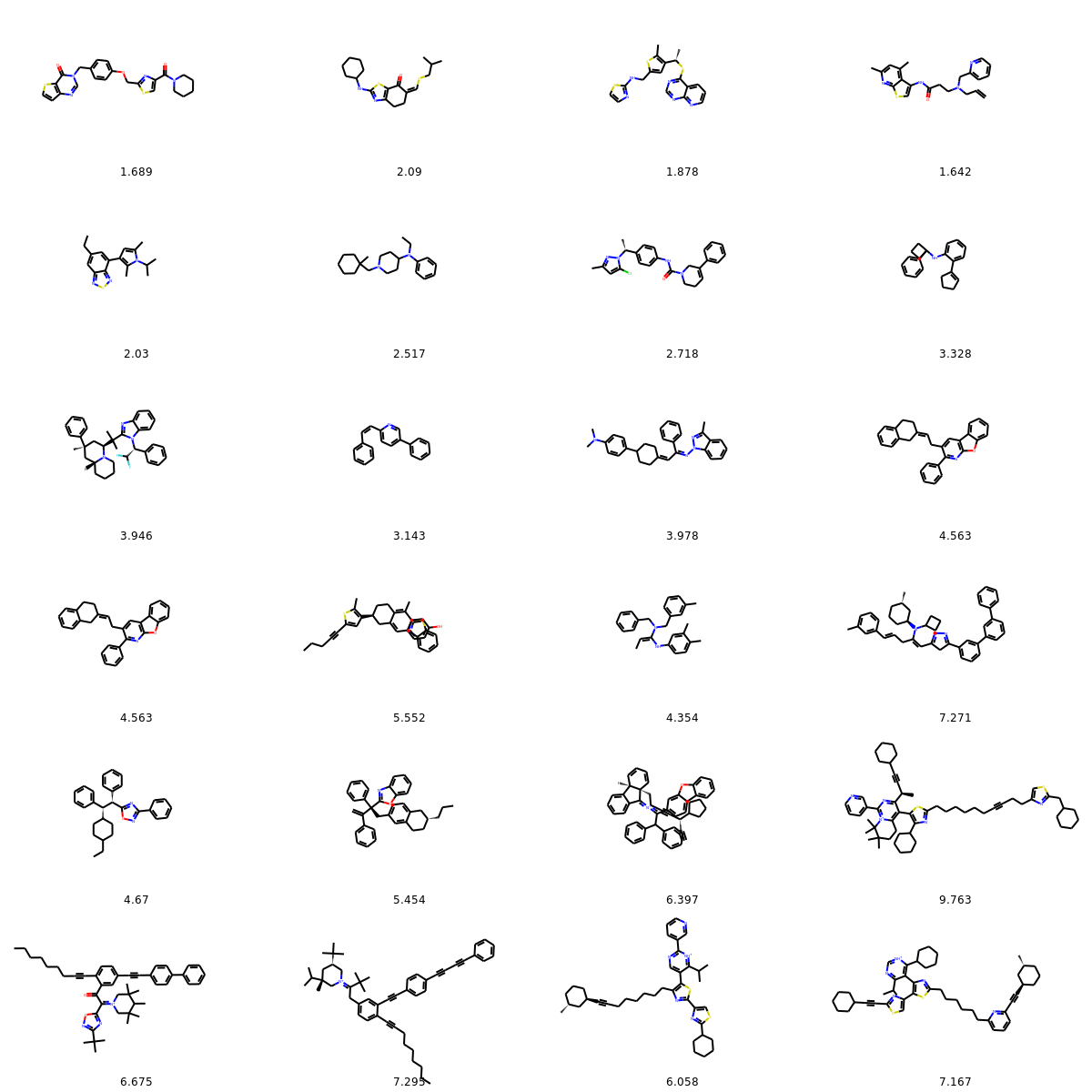}\label{fig:optimization_iters_best_mols_1}
\caption{Molecules with the highest penalized logP in the set being optimized for iterations 1-24 for unconstrained optimization. We show values of penalized logP below the molecules.} 
\label{figC1}
\end{figure}

\section{Conclusions} \label{sec:conclusions}
In this work, we introduce Mol-CycleGAN -- a new model based on CycleGAN which can be used for the \textit{de novo} generation of molecules. The advantage of the proposed model is the ability to learn transformation rules from the \textit{sets} of compounds with desired and undesired values of the considered property. The model operates in the latent space trained by another model -- in our work we use the latent space of JT-VAE. The model can generate molecules with desired properties, as shown on the example of structural and physicochemical properties. The generated molecules are close to the starting ones and the degree of similarity can be controlled via a hyperparameter. In the task of constrained optimization of drug-like molecules our model significantly outperforms previous results. In the future work we plan to extend the approach to multi-parameter optimization of molecules using StarGAN.\cite{choi2017stargan} It would also be interesting to test the model on cases where a small structural change leads to a drastic change in the property (e.g. the so-called activity cliffs) which are hard to model.

\begin{suppinfo}
All code used to produce the reported results can be found online at \url{https://github.com/ardigen/mol-cycle-gan}.
\end{suppinfo}

\begin{acknowledgement}
We would like to thank Sabina Podlewska for her helpful comments and for fruitful discussions.
\end{acknowledgement}

\bibliography{MolCycleGan}

\end{document}